\documentclass[runningheads]{llncs}
\usepackage[T1]{fontenc}
\usepackage{graphicx,verbatim}
\usepackage{amsmath,amssymb,amsfonts}
\usepackage{xcolor}
\usepackage{array}
\begin{document}
\title{3D Acetabular Surface Reconstruction from 2D Pre-operative X-ray Images using SRVF Elastic Registration and Deformation Graph}

\author{Shuai Zhang\inst{1,2} \and
Jinliang Wang\inst{3} \and
Sujith Konandetails \inst{4} \and
Xu Wang\inst{1,5} \and
Danail Stoyanov\inst{1,2} \and
Evangelos B.Mazomenos\inst{1,5}
\authorrunning{F. Author et al.}
%
\institute{The UCL Hawkes Institute, University College London, London, UK \and
The Department of Computer Science, University College London, London, UK \and
Joint Disease Department,Zhengzhou Orthopaedic Hospital, Zhengzhou, China
\and
University College Hospital, London, UK \and
The Department of Medical Physics and Biomedical Engineering, University College London, London, UK \\
\email{shuai.z@ucl.ac.uk}}}

\maketitle              
\begin{abstract}

Accurate and reliable selection of the appropriate acetabular cup size is crucial for restoring joint biomechanics in total hip arthroplasty (THA). This paper proposes a novel framework that integrates square-root velocity function (SRVF)-based elastic shape registration technique with an embedded deformation (ED) graph approach to reconstruct the 3D articular surface of the acetabulum by fusing multiple views of 2D pre-operative pelvic X-ray images and a hemispherical surface model.
The SRVF-based elastic registration establishes 2D-3D correspondences between the parametric hemispherical model and X-ray images,
and the ED framework incorporates the SRVF-derived correspondences as constraints to optimize the 3D acetabular surface reconstruction using nonlinear least-squares optimization.
Validations using both simulation and real patient datasets are performed to demonstrate the robustness
and the potential clinical value of the proposed algorithm. The reconstruction result can assist surgeons in selecting the correct acetabular cup on the first attempt in primary THA, minimising the need for revision surgery. Code and data will be released upon acceptance.


\end{abstract}
\section{Introduction}

Total hip arthroplasty (THA) is an effective procedure for treating end-stage hip arthritis, involving the replacement of hip joint with a prosthetic implant.
The number of primary THA has been rising worldwide due to growing prevalence of hip joint arthritis and increased access to orthopaedic care. 
Statistics indicate that
annual primary THA cases in the USA increased from 262K in 2019 to a projected 719K by 2040 \cite{shichman2023projections}, while the UK reported 94,936 procedures in 2018, with a predicted 40\% increase by 2060 \cite{matharu2022projections}.

Precise acetabular cup sizing is critical in THA, as it influences initial implant stability, long-term survivorship, and restoration of hip biomechanics~\cite{krishnamoorthy2015accuracy,feng2022primary,karampinas2024technical}. Both over-sizing and under-sizing pose significant risks: over-sizing may lead to acetabular fracture or impingement, whereas under-sizing compromises fixation and increases aseptic failure \cite{karampinas2024technical}.
Selecting the precise size can be challenging, particularly for less experienced surgeons.
Errors in sizing contribute to postoperative complications requiring revision THA, which is projected to rise significantly. Annual revision THA cases in the USA are estimated to increase by 78\%–182\% from 2014 to 2030, potentially reaching 572K by 2030 \cite{bains2024epidemiology}.

To pre-operatively predict the acetabular cup size, 2D digital templating remains the standard in clinical practice~\cite{bishi2022comparison}, achieving exact size prediction in approximately $70\%$ of cases \cite{krishnamoorthy2015accuracy}. Computer software is used to template the implant size and position on digital radiographs obtained from  pelvic X-ray imaging~\cite{shon2016acetabular}.
However, 2D templating exhibits inherent limitations in representing 3D osseous anatomy and is prone to magnification and patient positioning errors \cite{shaarani2013accuracy,brenneis2021accuracy}, which causes challenges in selecting the correct size. Surgeons can also assess the size of the acetabular cup intra-operatively \cite{karampinas2024technical}, with a trial-and-error approach. 
Less experienced surgeons are more prone imprecise selections causing patient pain and necessitating follow-up surgeries \cite{kenney2019systematic}.
This trial-and-error process becomes more challenging when selecting sizes for both the acetabular and femoral components, due to their interdependent sizing requirements \cite{karampinas2024technical}.

Advancements in CT-based 3D pre-operative planning demonstrate superior accuracy: $99.7\%$ exact size prediction in 290 cases compared to 2D templating \cite{crutcher2024comparison}. 
While CT imaging enables precise quantification of bone defects and implant sizing, reconstructing 3D acetabular surface models from multiple pre-operative 2D pelvic X-rays offers a promising alternative to reduce reliance on CT.






Prior works \cite{zheng2006reconstruction,zheng20092d,baka20112d,pan20233d,gu20243ddx} on 3D bone surface reconstruction using 3D statistical shape models (SSMs) and 2D X-ray images with paired CT models through 2D-3D registration, achieve sub-millimeter accuracy in reconstructing bone structures (e.g., femoral heads, tibia, pelvis).
However, SSMs requiring large volumes of annotated training data and Euclidean distance-based registration methods (e.g., iterative closest point (ICP) variants \cite{besl1992method}) frequently fail due to projection ambiguities and scale$/$shape discrepancies between models and images, making these methods less effective for acetabular reconstruction.

Accurate 3D acetabular surface reconstruction from pre-operative X-ray images holds significant potential for implant selection in THA. 
In this paper, we propose a robust and accurate 3D acetabular surface reconstruction method using minimum three views of 2D pre-operative pelvic X-ray images and a standard hemispherical surface model. This model approximates the acetabular morphology and is iteratively deformed using an integration of square-root velocity function (SRVF)-based elastic registration module and an embedded deformation (ED) graph \cite{sumner2007embedded}, aligning the projected contours with expert-annotated acetabular rim contours on the X-rays through a nonlinear least-squares optimization. The reconstructed 3D acetabular surface model can support surgeons in determining the optimal acetabular cup sizing, thereby enhancing surgical accuracy and patient outcomes. The proposed method uses routine pre-operative X-ray imaging and 
can be seamlessly integrated into clinical practice 
reducing the dependence on CT for 3D acetabulum modeling, thus minimizing radiation exposure and associated costs.

\section{Problem Description}
\begin{figure*}[!t]
	\centering
	\includegraphics[width=1.0\textwidth]{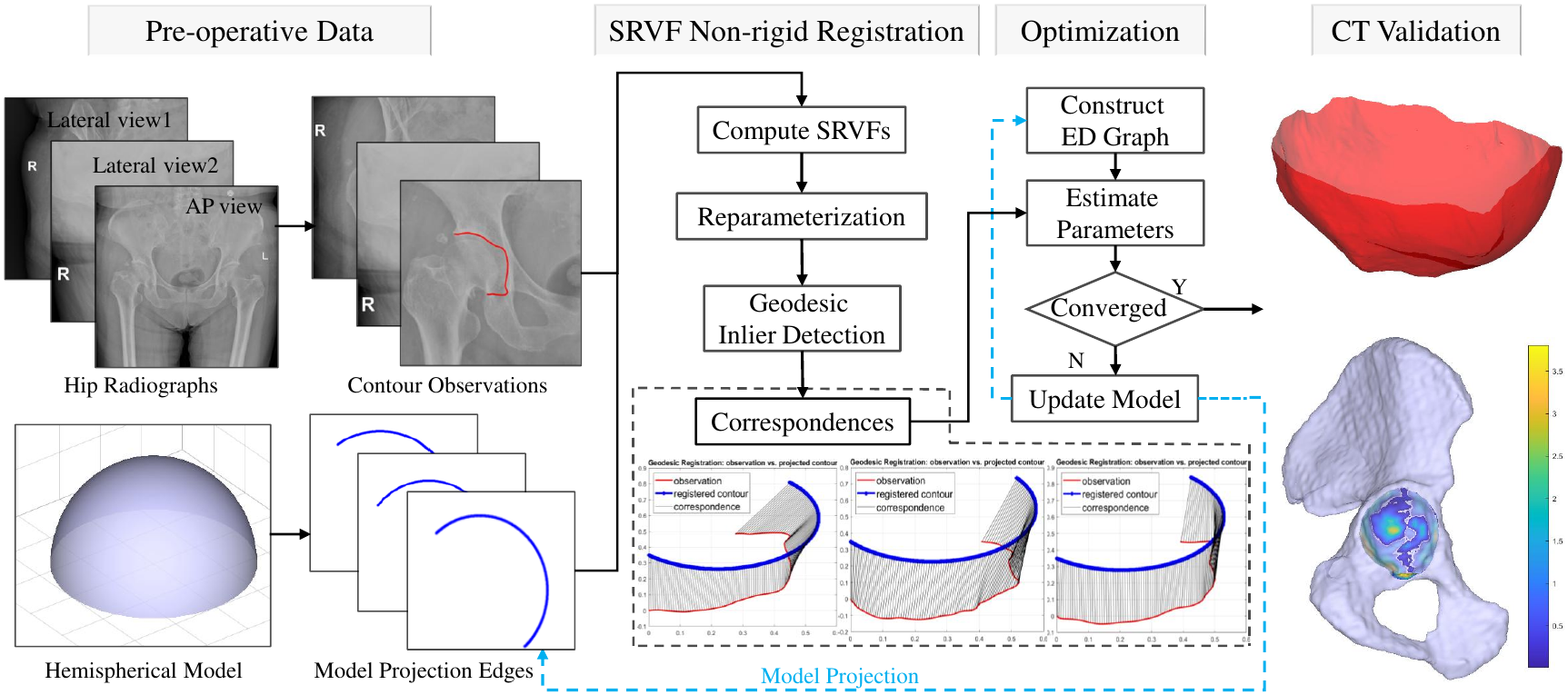}
	\caption{Main processes of the proposed acetabular surface reconstruction framework.}
	\label{fig:framework}
\end{figure*}
Fig.~\ref{fig:framework} shows the proposed framework for reconstructing a patient-specific 3D acetabular surface using three views of 2D pre-operative pelvic X-rays and a 3D hemispherical model. It includes pre-operative data preparation, SRVF non-rigid registration, ED graph optimization and validation using pelvic CT scans. 


Suppose $M$ represents the hemispherical surface model consisting of a set of 3D vertices $\{ {\mathbf{P}_i} \}$, $i \in \{ 1, ..., N_P \}$ and $N$ views of pre-operative pelvic X-ray images are captured. Let $\mathcal{C}_{k}=\{\mathbf{p}_{k,1},...,\mathbf{p}_{k,N_{k}}\}$ represent the contour observations (red curves in Fig.~\ref{fig:framework}) of the acetabulum on the $k$-th X-ray image, $k \in \{1, ..., N\}$ and $N_{k}$ is the number of acetabular bone contour pixels extracted from the $k$-th X-ray image.
To estimate the optimal ED parameters that can deform model $M$ into a patient-specific acetabulum model with the updated positions of all vertices, such that the 2D projections from the updated model match with the pre-operative X-ray observations, model vertices are projected onto the $N$ views of X-ray images using the perspective projection model of X-ray fluoroscopy:
\begin{equation}\label{Eq_Projection}
	\begin{aligned}\
		\phi(\mathbf{P}_{i}, R_k, \mathbf{t}_k) = [\mathbf{p}_{i}^T, 1]^T\propto K (R_k \mathbf{P}_{i} + \mathbf{t}_k),
	\end{aligned}
\end{equation}
where $\mathbf{p}_{i} \in \mathbb{R}^2$ is the projection of vertex $\mathbf{P}_{i}$ on the $k$-th image, and $K$ is the X-ray camera intrinsic matrix. $\{R_k \in SO{(3)}, \mathbf{t}_k \in \mathbb{R}^3\}$ represent the rotation matrix and translation vector that transform vertices from the model frame to the $k$-th X-ray frame, which are assumed to be known when estimating ED parameters.
The alpha-shape method \cite{kirkpatrick1983shape} is used to extract the model projection edge points $\tilde{\mathcal{C}}_{k} = \{\tilde{\mathbf{p}}_{k,1},..., \tilde{\mathbf{p}}_{k,N_k^M}\}$, which are shown as blue curves in Fig.~\ref{fig:framework}.

Thus, the problem considered in this paper is,
given $N$ views of 2D pre-operative X-ray observations and the 3D hemispherical surface model $M$, how to estimate the optimal deformation graph, then the actual acetabular surface can be obtained by deforming the standard hemispherical surface model using the ED graph, where the poses of $N$ views of X-ray images assumed to be known.


\section{Methodology}\label{Sec:Methodology}


\subsubsection{Preliminaries of Embedded Deformation:}

The embedded deformation (ED) graph is used to manipulate the shape deformation and achieve 3D acetabulum surface reconstruction. It consists of a set of nodes $\{\mathbf{g}_j  \in \mathbb{R}^3\}, j \in \{1, ..., N_{E}\}$ down sampled uniformly from the input 3D model, each associated with an affine transformation $\{A_j \in \mathbb{R}^{3 \times 3}, \mathbf{t}_j \in \mathbb{R}^3 \}$, and influences nearby vertices through weighted interpolation. For a model vertex $\mathbf{P}_i$, its position after deformation is calculated using its $m$ neighboring nodes:
\begin{equation}\label{Eq:ED_manipulation}
\tilde{\mathbf{P}}_i = \sum_{j=1}^{m}w_{ij} \left(A_j(\mathbf{P}_i - \mathbf{g}_j)+ \mathbf{g}_j+ \mathbf{t}_j \right), w_{ij} = 1- ||\mathbf{P}_i - \mathbf{g}_j||/d_{max},
\end{equation}
where $d_{max}$ is the distance between $\mathbf{P}_i$ and the $m+1$ nearest node.

\subsubsection{SRVF Based 2D-to-3D Non-rigid Registration:}
The calculation of 2D-3D correspondences between the model vertices and observations is simplified to the alignment of 2D curves between the model projection edges and observation contours.
By using the square-root velocity function (SRVF) representation, the problem of 2D curves alignment is transformed into an optimization problem that is invariant to rotation, translation and scale.

The curves $\mathcal{C}_k$ and $\tilde{\mathcal{C}}_{k}$ are transformed to their SRVF representations, $\mathbf{q}_{\mathcal{C}_k}$ and $\mathbf{q}_{\tilde{\mathcal{C}}_{k}}$. 
Then, the alignment problem is formulated as estimating the optimal rotation $O \in SO(2)$ and reparameterization function $\gamma$ that minimizes the $\mathbb{L}_2$ distance (a proxy for geodesic distance) between the two SRVFs: 
\begin{equation}\label{Eq:SRVFs}
	 \underset{O, \gamma}{\mbox{min}}  \left\|O (\mathbf{q}_{\tilde{\mathcal{C}}_{k}}(\gamma))\sqrt{\dot{\gamma}}-\mathbf{q}_{\mathcal{C}_k} \right\|^{2}
\end{equation}
where $\gamma$ is a diffeomorphic warping function. Dynamic programming is used to solve the optimization problem of Eq.(\ref{Eq:SRVFs}). The resulting aligned curves are then reconstructed from the optimized SRVFs, and the correspondences between the 3D model vertices and the 2D observations can be inferred and used to provide constraints for optimizing ED parameters. It is noted that, unlike Euclidean 2D-3D registration methods that recalculate correspondences in each iteration, SRVF's mathematical robustness (invariance to parameterization, rotation, and scaling) allows only a few times of correspondences calculation during the optimization ($1$ to $3$ times in our experiments).

\subsubsection{Reconstruction Formulation and Optimization:}
The ED deformation parameters of affine transformation are estimated by using:
\begin{equation}\label{Eq:optimization}
	\underset{\{A_j, \mathbf{t}_j \}_{j=1}^{N_E}} {\mbox{min}} w_{rot}E_{rot} + w_{reg}E_{reg} + w_{obs}E_{obs},
\end{equation}

which consists of the three terms $E_{rot}$, $E_{reg}$ $E_{obs}$ with
weights $w_{rot}$, $w_{reg}$, $w_{obs}$, respectively.
The rotation term $E_{rot}$ is used to make the deformation as rigid as possible by constraining the affine matrices close to rotations, while the regularization term $E_{reg}$ is used to ensure a smooth deformation by preventing divergence of the neighboring nodes, for details refer to \cite{sumner2007embedded}. $E_{obs}$ is the new defined observation term penalizing the misalignment between the observed contours in the pre-operative X-ray images and contours extracted from the projection of the deformed model obtained in Eq.(\ref{Eq:ED_manipulation}):
\begin{equation}
	E_{obs} = \sum_{k=1}^{N} dist(\mathcal{C}_k, \tilde{\mathcal{C}}_{k}) = \sum_{k=1}^{N} \sum_{i \in \mathbb{N}(k)} \left\| \phi(\tilde{\mathbf{P}}_{k_i}, R_k, \mathbf{t}_k) - \mathbf{q}_{k,i}\right\|^{2},
\end{equation}
where $\mathbb{N}(k)$ is the index set of observations in the k-th X-ray image that have corresponding 3D model vertices, and $\tilde{\mathbf{P}}_{k_i}$ is calculated from $\mathbf{P}_{k_i}$ by using Eq.(\ref{Eq:ED_manipulation}).

The Gauss-Newton (GN) method is used to solve the formulated non-linear optimization problem in Eq.(\ref{Eq:optimization}). The optimization uses a two-stage strategy to balance efficiency and accuracy: initial correspondence calculation and adaptive correspondence refinement.
First, 2D-3D correspondences are calculated in the first iteration and fixed for the following iterations until the deformation changes fall below a threshold. This generates an anatomically accurate deformation approximating the target acetabular shape. Then, once the deformation stabilizes, new correspondences are calculated to refine local geometric mismatches (e.g., acetabular rim curvature or fossa details). In experiments, correspondence calculation are triggered 1 to 3 times in total.
Besides, at the start of each iteration, the ED graph is reinitialized to ensure that the topology updates and transformations are consistently derived from the latest deformed shape, preventing errors accumulation and enhancing the robustness and accuracy of the optimization.

\section{Experiments}
We validate our 3D acetabular surface reconstruction framework using simulated and real patient data and compare against classical ICP \cite{besl1992method} which uses closest Euclidean distance to calculate correspondences, and the feature-based Euclidean 2D-3D registration method (ICP-NormVec) \cite{pan20233d} that combines nearest neighbor matching with normal vector angular thresholding for outliers rejection.
The pelvis reconstruction work \cite{gu20243ddx} which combines X-ray depth estimation networks with SSMs is not compared, due to the unavailability of their code and large-scale annotated dataset.
Additionally, our limited dataset (five pelvis CT scans and a few number of X-rays) is insufficient for constructing reliable SSMs.

Due to the narrow angular separation among different views of pelvic X-ray images and severe occlusion from the pelvis, spine, and surrounding bones makes critical anatomical features (e.g., roof, teardrop) unclear, the minimum number of X-ray images used in simulations and real experiments is three, then $N = 3$.


\begin{figure*}[htbp!]
	\centering
	\includegraphics[width=1.0\textwidth]{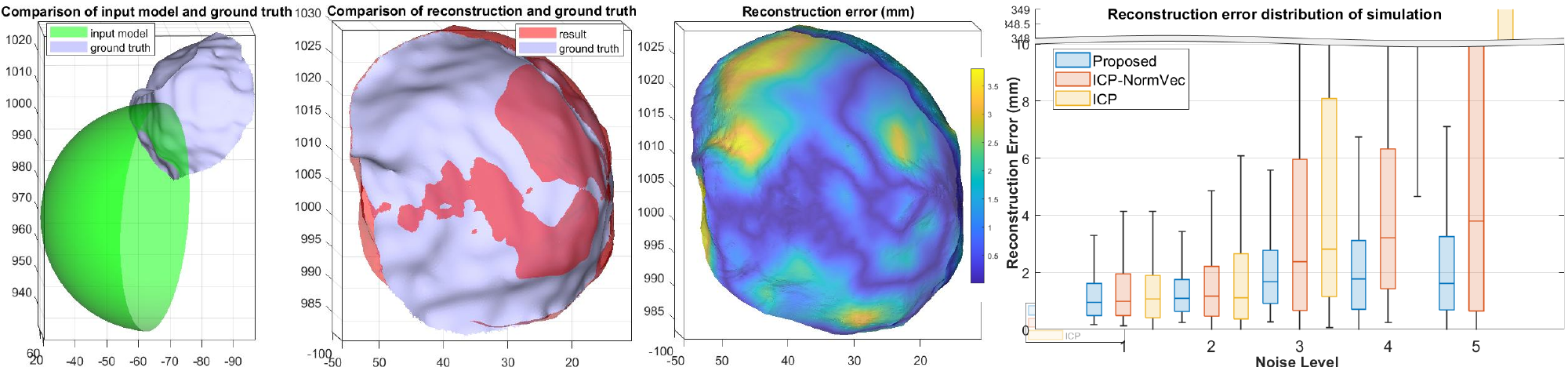}
	\caption{Reconstruction results from simulation experiments: the 1st, 2nd and 3rd columns show the results for an example at noise level $5$: the standard hemispherical surface model (green) and the CT-segmented acetabular surface model (grey), the comparison between the reconstruction (red) and the CT-segmented acetabular surface model, and the reconstruction error, respectively; the 4th column presents the error distribution for simulated data with the increasing noise levels 1 to 5.}
	\label{fig:simulation_fig}
\end{figure*}

\subsubsection{Simulated Data Experiments:}
To simulate the real scenario, one anteroposterior (AP) and two lateral views are generated from one CT-segmented acetabular surface model by setting the rotation angles (around the body's longitudinal axis) as $\{0^\circ, 20^\circ, -20^\circ\}$.
Zero-mean Gaussian noise with standard deviation
(SD) of 2 pixels is added to the contour feature observations on the simulated X-ray images.
Then, the scale, position, and pose of the standard hemispherical surface model are initialized by fitting it to the CT-segmented acetabular surface model.

Five different levels of noise are added to the rotation, translation, and scale of the standard hemispherical surface model and used in the 3D reconstruction simulation. Zero-mean Gaussian noise with SD of
$\{0.1, 0.2, 0.3, 0.4, 0.5\}$ rad is added to the rotation Euler angles, noise with SD of
$\{10, 30, 50, 80, 100\}$ mm is added to the translations, and noise with SD of $\{1.1, 1.2, 1.3, 1.4, 1.5\}$ is added to the scale of the hemispherical surface model.
Ten independent runs are executed for each noise level to test the robustness
and accuracy of the proposed algorithm against compared methods.

Fig.~\ref{fig:simulation_fig} shows one example of 3D reconstruction result on the $5$-th level of noise, and the reconstruction error distribution for simulated data with increasing noise levels $1$ to $5$. The majority of the acetabular surface have been recovered with low error (blue color in 3rd column). The reconstruction error distribution (compared to the CT segmented acetabular surface model) of the ten runs is presented in the $4$-th column, it shows that the proposed framework has the highest accuracy and robustness in terms of mean absolute errors (MAEs) and SD, and for all the methods, when the noise level is very small, the reconstruction accuracy is close to the original result. However, as the levels
of noise for rotation, translation and scale of the input model increase simultaneously, 
only the proposed method maintains accurate 3D reconstruction with small MAEs and SD, proving its robustness and high accuracy.
The main reasons are: i) the proposed SRVF elastic registration approach is able to provide robust and accurate
correspondences; ii) the regularization term in Eq.(\ref{Eq:optimization}) designed for
smoothing the deformation contributes to the robustness under noisy observations; and iii) the deformation graph is reinitialized based on the latest shape, preventing error accumulation.

\begin{figure*}[!t]
	\centering
	\includegraphics[width=1.0\textwidth]{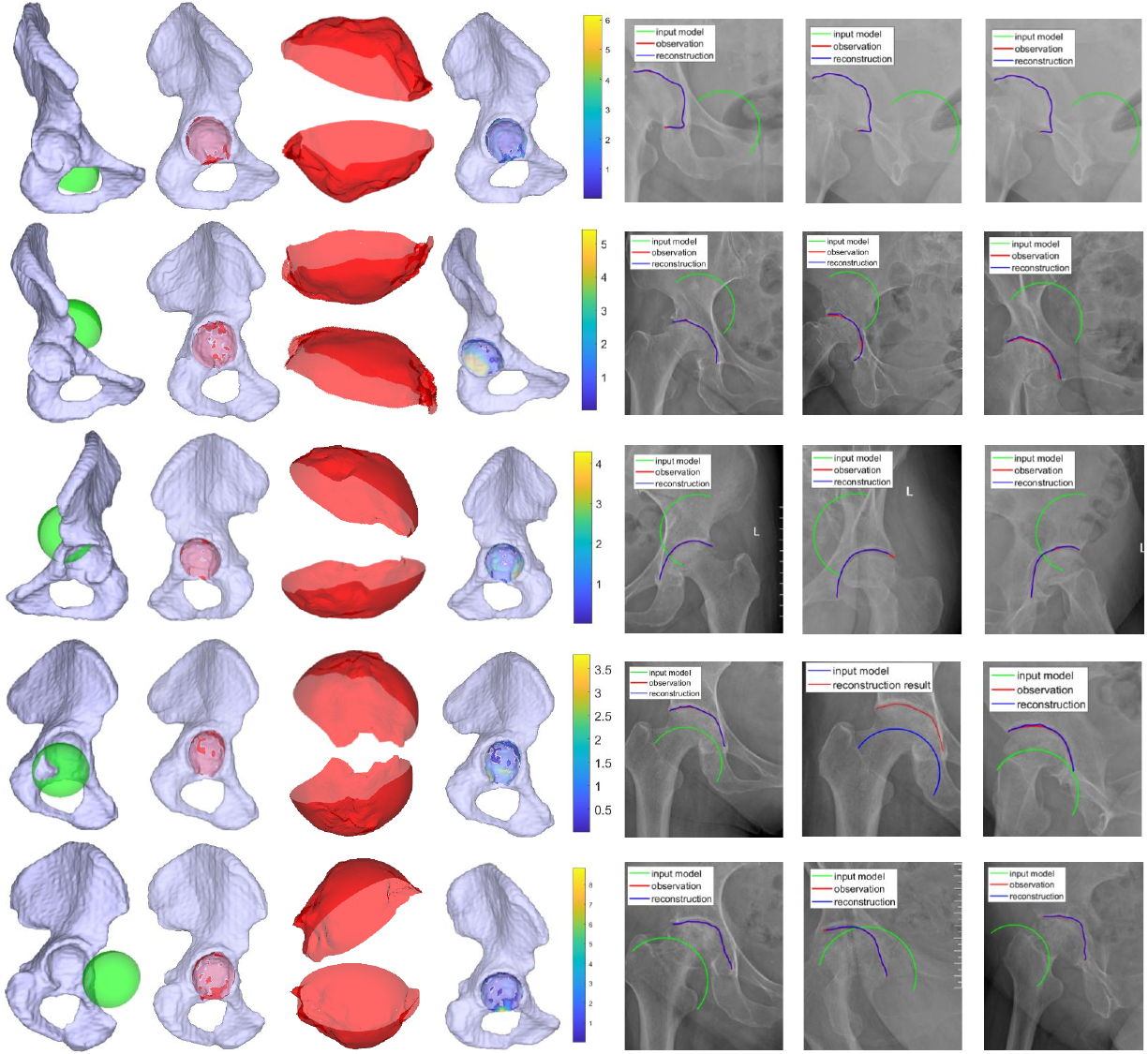}
	\caption{Reconstruction result from patient data using three pre-operative X-ray images. Each row shows the experimental result on one patient data. The 1st column shows the hemispherical surface model (green) and CT-segmented pelvis model (grey). The 2nd column shows the reconstruction result (red) compared to the corresponding pelvis model. The 3rd column shows the reconstruction in two different viewing angles. The 4th column shows the reconstruction error. The last three columns show the three pre-operative X-ray images. The projection contours of the input hemispherical model (green) and reconstruction result (blue), and the observation (red) are also presented.}
	\label{fig:real_exp}
\end{figure*}

\subsubsection{Validation using patient data:}
For clinical validation, datasets from five primary THA patients were collected, each including pre-operative pelvic CT scans, AP and two lateral views of pre-operative pelvic X-ray images, and intra-operatively confirmed cup dimensions. 
The X-rays (AP and two lateral views) were captured by maintaining the pelvis stable (the same supine position) and maximizing the angular disparity.
Acetabular contour observations were independently annotated by two orthopaedic surgeons to ensure reliability.
The anterior and posterior walls of the acetabulum are often superimposed with other osseous structures (e.g., ilium, ischium) or soft tissues (e.g., intestinal gas, fat), thus blurry edges of the walls are not used as observation in the real experiments. The poses of X-ray images used in our reconstruction validation are obtained from 2D-3D registration to CT-segmented pelvic models.

Fig.~\ref{fig:real_exp} shows the reconstruction result obtained from the five real patient data. The 1st column shows the difference in pose and scale between the acetabular regions of the CT-segmented pelvis models and the standard hemispherical surface model. 
The error map in the 4th column indicates that most areas of the acetabular surface are reconstructed with low error (blue).
After reconstruction, the projection contours of the reconstructed model (blue curves) are almost the same as the observations (red curves) from the pre-operative X-ray images.
The MAEs with SD of the 3D reconstruction (compared to the CT segmented pelvis models) are reported in Table \ref{tab:real-exp-table-one}, where we can see that the MAEs and SD of the 3D reconstruction are around 0.98 $mm$ to 1.3 $mm$ and 0.95 $mm$ to 1.25 $mm$, respectively. In contrast, the compared methods using the open curve observations exhibit significantly larger errors due to the algorithm instability.

Based on the reconstructed acetabular surface models, the acetabular cup size is estimated by fitting hemispherical surface models, with the diameter of the fitted models used as the cup size. Compared to the intra-operatively confirmed cup size, the estimation error is 1.3 $mm$, with a standard deviation of 1.26 $mm$.

\begin{table}[!t]
	\centering
	\caption{Summary of the mean absolute errors and SD of the reconstruction results on the five real patient data.}
	\label{tab:real-exp-table-one}
	\begin{tabular}{|c|cc|cc|cc|cc|cc|}
		\hline
		PatientID                           & \multicolumn{2}{c|}{1}                                          & \multicolumn{2}{c|}{2}                                          & \multicolumn{2}{c|}{3}                                          & \multicolumn{2}{c|}{4}                                         & \multicolumn{2}{c|}{5}                                          \\ \hline
		Metrics($mm$)                            & \multicolumn{1}{c|}{MAEs}          & SD                        & \multicolumn{1}{c|}{MAEs}          & SD                        & \multicolumn{1}{c|}{MAEs}          & SD                        & \multicolumn{1}{c|}{MAEs}          & SD                       & \multicolumn{1}{c|}{MAEs}          & SD                        \\ \hline
		\multicolumn{1}{|l|}{Initial Error} & \multicolumn{1}{l|}{40.04}         & \multicolumn{1}{l|}{16.74} & \multicolumn{1}{l|}{33.13}         & \multicolumn{1}{l|}{10.07} & \multicolumn{1}{l|}{19.20}         & \multicolumn{1}{l|}{12.11} & \multicolumn{1}{l|}{9.27}          & \multicolumn{1}{l|}{7.04} & \multicolumn{1}{l|}{14.13}         & \multicolumn{1}{l|}{10.95} \\ \hline
		Proposed                            & \multicolumn{1}{c|}{\textbf{1.40}} & \textbf{1.22}              & \multicolumn{1}{c|}{\textbf{1.63}} & \textbf{1.61}              & \multicolumn{1}{c|}{\textbf{1.19}} & \textbf{0.95}              & \multicolumn{1}{c|}{\textbf{1.34}} & \textbf{1.25}             & \multicolumn{1}{c|}{\textbf{0.98}} & \textbf{0.98}              \\ \hline
		ICP-NormVec                         & \multicolumn{1}{c|}{2.61}          & 2.65                       & \multicolumn{1}{c|}{32.90}         & 130.9                      & \multicolumn{1}{c|}{23.76}         & 6.64                       & \multicolumn{1}{c|}{13.27}         & 12.30                     & \multicolumn{1}{c|}{13.32}         & 11.17                      \\ \hline
		ICP                                 & \multicolumn{1}{c|}{119.0}         & 13.45                      & \multicolumn{1}{c|}{257.3}         & 197.8                      & \multicolumn{1}{c|}{234.4}         & 187.8                      & \multicolumn{1}{c|}{30.64}         & 26.63                     & \multicolumn{1}{c|}{2.64}          & 2.41                       \\ \hline
	\end{tabular}
\end{table}

\section{Conclusion}
This paper presents a novel method for 3D acetabulum reconstruction using pre-operative pelvic radiographs and a standard hemisphere model.
Simulation experiments demonstrate the robustness and accuracy of the proposed algorithm. Real patient data experiments demonstrate the feasibility and potential clinical value of the proposed framework. In conclusion, the algorithm proposed in this paper can be deployed to support acetabular cup size prediction pre-operatively without the changing on the pre-operative primary THA procedures. In the future, we aim to improve this work to reconstruct the patient-specific 3D acetabular surface map using the pre-operative AP view X-ray image only combined with synthesized views of pelvic X-ray images.

%
%
\bibliographystyle{splncs04}
\bibliography{mybibliography}

%
%
%
%
%

\end{document}